\pdfoutput=1
\documentclass[11pt, a4paper, nonumbering]{paper}

\usepackage[T1]{fontenc}
\usepackage[utf8]{inputenc}
\usepackage{graphicx}
\usepackage{booktabs}
\usepackage{array}
\usepackage{multirow}
\usepackage{amsmath}
\usepackage{amssymb}
\usepackage{xspace}
\usepackage{enumitem}
\usepackage{caption}
\usepackage[table]{xcolor}
\usepackage{colortbl}
\usepackage{titlesec}
\usepackage[authoryear, sort&compress, round]{natbib}

\definecolor{wcaClaude}{HTML}{2A78D6}   
\definecolor{wcaGPT}   {HTML}{EB6834}   
\definecolor{wcaGemini}{HTML}{1BAF7A}   
\definecolor{wcaKimi}  {HTML}{EDA100}   
\definecolor{wcaGLM}   {HTML}{E87BA4}   
\definecolor{wcaSeed}  {HTML}{008300}   
\definecolor{wcaBase}  {HTML}{4A3AA7}   
\definecolor{wcaTruth} {HTML}{D03B3B}   
\definecolor{wcaHead}  {HTML}{184F95}   
\definecolor{wcaTint}  {HTML}{EDF3FD}   
\definecolor{wcaTintB} {HTML}{F5F4F1}   
\definecolor{wcaRule}  {HTML}{9FB4CE}
\definecolor{wcaInk2}  {HTML}{52514E}

\hypersetup{colorlinks=true,linkcolor=wcaHead,citecolor=wcaHead,urlcolor=wcaHead}
\usepackage[capitalize,noabbrev]{cleveref}

\titleformat{\section}{\large\bfseries\headingfont\color{wcaHead}}{\color{wcaHead}\thesection.}{0.5em}{#1}[]
\titleformat{name=\section,numberless}{\large\bfseries\headingfont\color{wcaHead}}{}{0em}{#1}[]
\titleformat{\subsection}{\bfseries\color{wcaHead}}{\color{wcaHead}\thesubsection.}{0.5em}{#1}[]
\titleformat{\subsubsection}{\bfseries\itshape\color{wcaHead}}{\color{wcaHead}\thesubsubsection.}{0.5em}{#1}[]
\titleformat{\paragraph}[runin]{\bfseries\color{wcaHead}}{}{0em}{#1}

\newcommand{\thead}{\rowcolor{wcaTint}\color{wcaHead}}
\newcommand{\mdl}[2]{\textcolor{#1}{$\bullet$}\,#2}

\newcommand{\bench}{\textsc{WorldCup Arena}\xspace}

\newcommand{\sbench}{\textsc{SocietyBench}\xspace}

\title{\bench: Prospective, Leakage-Free Evaluation of
  Frontier LLMs on a Live Tournament}
\renewcommand{\runningtitle}{\bench: forecasting a live World Cup}
\reportnumber{}

\newsavebox{\wcafront}
\AtBeginDocument{\savebox{\wcafront}{%
  \setlength{\fboxsep}{6pt}%
  \colorbox{wcaTint}{%
  \begin{tabular}{@{}r@{\ \ }l@{}}
  \small\textcolor{wcaHead}{\textbf{Project page}} &
    \small\href{https://co-minder.github.io/worldcup2026/}{\texttt{co-minder.github.io/worldcup2026}}
  \end{tabular}}%
}}

\reportnumber{}
\correspondingauthor={Equal contribution.\\
Correspondence: \href{mailto:qizhangyang2000@gmail.com}{qizhangyang2000@gmail.com}, \href{mailto:zhenran.w.1103@gmail.com}{zhenran.w.1103@gmail.com}}

\author[*]{%
  Zhenran Wang\textsuperscript{*} \quad Zhonghan Bian\textsuperscript{*} \quad Jinsong Li \quad Zhangyang Qi
  \par\vspace{11pt}\centerline{\usebox{\wcafront}}%
}

\begin{abstract}
\noindent
Benchmarks that measure the forecasting ability of large language models are almost always
retrospective: the event has happened, the answer is somewhere on the Web, and the
evaluation must defend itself against memorisation. We report the opposite design. Over the
39 days of the 2026 FIFA World Cup, six frontier LLMs---all with extended thinking and
native server-side web search---were asked before every kickoff, one match at a time, to
fill in a seven-market prediction card for all 104 matches, plus 12 group winners and a
pre-tournament outright pool; no answer existed when the question was asked, so the
evaluation is leakage-free by construction rather than by filtering, and the frozen archive
holds 4{,}494 scored predictions. What the tournament establishes is a set of behaviours the six systems share. On match
outcome they average 63.9\%, level with backing the bookmaker's favourite---which is in fact
what they usually do. They agree with one another far more often than they are right, so a
majority vote adds nothing. They under-commit to draws and to goals, and crowd their
scoreline picks onto a single prototypical result. Accuracy tracks how lopsided a fixture is
rather than how much is known about it: it collapses in the closest ties, where the dossiers
are richest, while questions about the tournament as a whole are answered well. On this task the current generation of frontier systems is not sharply differentiated: the standings hold up at the top and the bottom across the run and churn in the middle, and the margins stay narrow throughout. The briefing
dossiers, fixtures and official results are released as a benchmark, together with the
scoring code.\end{abstract}

\makeatletter
\renewcommand{\maketitle}{\bgroup\setlength{\parindent}{0pt}%
  \vspace*{3pt}%
  \begin{adjustwidth}{0pt}{0pt}
  \begin{center}
    {\titlefont \@title\par}%
    \vskip20pt
    {\@author\par}%
    \vskip30pt
  \end{center}
  \end{adjustwidth}
  \egroup
  {{\abscontent}}%
  \thispagestyle{firststyle}
}
\makeatother

\begin{document}
\maketitle

\section{Introduction}
\label{sec:intro}

Forecasting is an unusually honest test of a reasoning system. A model that has read the
entire Web can recall, summarise and re-derive; it cannot recall an event that has not
happened yet. This makes genuine future events the one evaluation setting in which
contamination is a category error rather than a threat to be mitigated: at the moment of
asking, the answer exists nowhere to be recalled.

Almost all published LLM forecasting evaluations nonetheless operate retrospectively.
Autocast~\citep{autocast} freezes questions from resolved tournaments;
LiveBench~\citep{livebench} and LiveCodeBench~\citep{livecodebench} rotate in newly
published items; detection work asks whether a benchmark has leaked at
all~\citep{sainz2023,timetravel}; our own \sbench~\citep{societybench} anonymises real
events into counterfactual worlds. Each is a proxy for the property one actually wants:
\emph{the answer did not exist when the model was asked}.
ForecastBench~\citep{forecastbench} attains it by continuously refreshing questions that
resolve in the future, and is the closest antecedent to this work.

\begin{figure}[t]
\centering
\includegraphics[width=\linewidth]{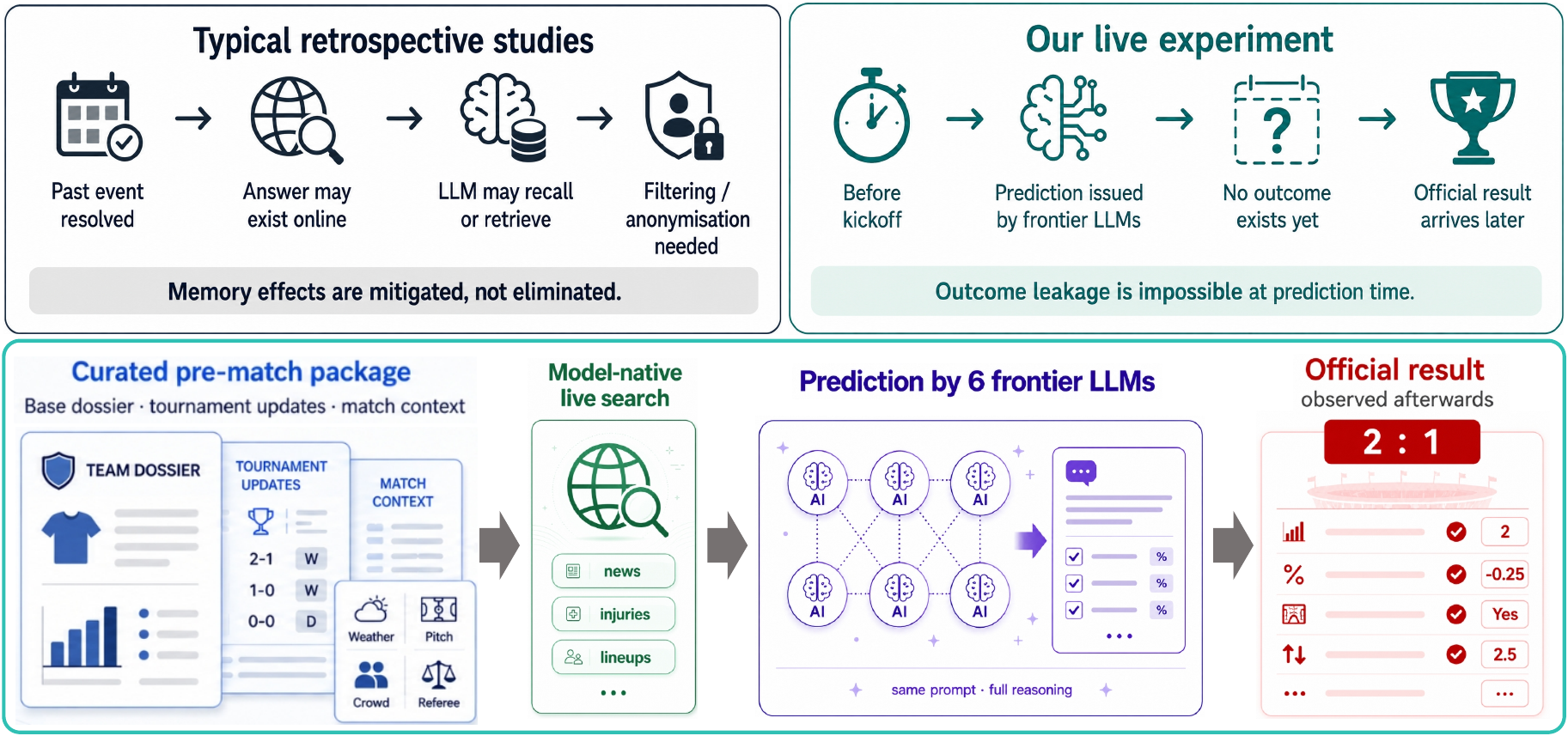}
\caption{\textbf{Top:} a retrospective evaluation asks after the event resolves; this study
asks before kickoff. \textbf{Bottom:} the pipeline at one fixture, from the frozen pre-match
package to the settled result.}
\label{fig:teaser}
\end{figure}

We report a study in which that property holds exactly, over a single dense, fully observed
event: the 2026 FIFA World Cup. Between 10~June and 19~July~2026 a daily pipeline refreshed
a dossier for each of the 48 teams, locked fixture, venue, kickoff, weather, referee and a
fixed bookmaker handicap line into a prompt header, and asked six frontier LLMs---all with
extended thinking and native web search---to commit to seven markets for the next match,
always before kickoff. All 104 were covered. Results were settled afterwards, so the input
to match $t{+}1$ contains the outcome of match $t$ and nothing later. \Cref{fig:teaser} sets the two designs
side by side and shows what happens at a single fixture.

Two design choices make the result harder to explain away. The models ran at full strength:
each used extended thinking at its highest available setting and the provider's \emph{own}
server-side web search, so at prediction time they could read the same morning's team news,
injury reports and odds that a human forecaster would. Every retrieval was executed by the
provider's own service, which keeps the comparison a test of the model's tool use, and 757
of the 848 stored responses carry a provider flag attesting that search was actually
exercised. Coverage is complete: all 104 fixtures, group stage through the final, every one
answered before kickoff and every run kept. Whatever the models did here, they did it with
live access to the world and on the complete fixture list.

The headline result is a leaderboard. The six span 897 to 813 points, and on this task the
current generation of frontier systems is not sharply differentiated: the top and bottom of
the table hold up across the run rather than emerging only at the end, the middle reorders
under a change of scoring, and the margins stay narrow throughout. The tournament also
resolves the behaviour the six models share.

They track the betting market rather than beat it, siding with the bookmaker's favourite on
85.6--90.4\% of matches for a mean accuracy, 63.9\%, that is within noise of the 64.4\%
obtained by mechanically backing that favourite. They are conservative in a way specific to
football, predicting 8--14 draws where 27 occurred and placing 28.0\% of all scoreline picks
on 2--1. They herd, agreeing on the outcome in 76.0\% of matches, so a majority vote adds
nothing. Their accuracy is a function of how lopsided the fixture is rather than of how much
is known---86.5\% in the round of 32 against 8.3\% in the semi-finals, where the dossiers
are richest. And they are macro-strong but micro-weak: the same systems answer
tournament-level questions 20 to 44 points better than single matches. \Cref{sec:markets}
gives the evidence.

\noindent\textbf{Contributions.} First, a fully prospective, leakage-free evaluation
protocol for LLM forecasting, executed end-to-end on a complete tournament with objective
ground truth. Second, a frozen public benchmark: the 48 team dossiers across 46 timestamped
snapshots, the 104-fixture list with team and venue metadata, and the official results that
answer it. Third, an uncertainty-aware reading of the leaderboard---bootstrap intervals and
five alternative scoring designs---together with five behavioural findings that survive all
of them.

\section{Related Work}
\label{sec:related}

\begin{table}[t]
\centering\small
\setlength{\tabcolsep}{5pt}
\begin{tabular}{@{}lll@{}}
\toprule
\thead
 & \sbench & \bench (this report) \\
\midrule
Temporal stance      & Retrospective            & \textbf{Prospective} \\
Leakage control      & 3-phase anonymisation    & \textbf{Answer did not exist} when asked \\
Ground truth         & Extracted from discourse & Official scoreline \\
Answer format        & Probability + date       & Single choice per market \\
Scoring axes         & Calibration, temporal    & Market-weighted points \\
Breadth              & 25 events, 2 languages   & 1 event, 104 matches, 39 days \\
Elicitation          & One-shot per cutoff      & Rolling, one match at a time \\
Key limitation       & Anonymisation is a proxy & Single event; one choice per market \\
\bottomrule
\end{tabular}
\caption{\bench is the prospective control condition for \sbench: it buys freedom from the
memory assumption at the cost of breadth.}
\label{tab:vs_societybench}
\end{table}

\paragraph{Agent benchmarks.} Most LLM and agent benchmarks target the
\textit{task-completion} axis: SWE-bench~\citep{swebench} resolves real GitHub issues,
WebArena~\citep{webarena} runs long-horizon web tasks, OSWorld~\citep{osworld} covers
desktop workflows, and GAIA~\citep{gaia} poses assistant questions requiring tool use. All
ask whether a system can execute a sequence of operations to reach a \textit{known} end
state, and all can be graded the moment the run finishes. \bench asks the opposite kind of
question: not whether a model can reach an answer that already exists, but whether it can
name one that does not exist yet.

\paragraph{Forecasting benchmarks for LLMs.} A separate line poses questions verifiable only
once events play out. Autocast~\citep{autocast} established the format of scoring LLMs
against resolved forecasting-tournament questions; Halawi et al.~\citep{halawi2024} show
that a retrieval-and-reasoning pipeline can approach human crowd accuracy; Schoenegger et
al.~\citep{schoenegger2023,silicon_crowd} run LLMs inside live tournaments and find that
ensembles rival crowds while individual models lag; ForecastBench~\citep{forecastbench}
makes the question refresh continuous, so that resolution always postdates the query. These
sample broadly and thinly---many questions, many domains, one shot each. \bench trades that
breadth for density: 104 tightly coupled predictions from a single domain, under a fixed
template, over 39 consecutive days, which is what makes within-tournament trends and
cross-model agreement measurable at all.

\begin{table}[t]
\centering\footnotesize
\setlength{\tabcolsep}{3.4pt}
\begin{tabular}{@{}llll@{}}
\toprule
\thead
Label & Model identifier & Thinking \emph{(highest available)} & Search \emph{(provider's own)} \\
\midrule
\mdl{wcaClaude}{Claude} & \texttt{claude-opus-4-8}            & thinking, 24k budget     & native (\texttt{web\_search})   \\
\mdl{wcaGPT}{GPT}    & \texttt{gpt-5.5}                    & reasoning effort high    & native (\texttt{web\_search})   \\
\mdl{wcaGemini}{Gemini} & \texttt{gemini-3.1-pro-preview}     & thinking budget 24{,}576 & native (\texttt{google\_search}) \\
\mdl{wcaKimi}{Kimi}   & \texttt{kimi-k2.6}                  & built-in thinking        & first-party (\texttt{\$web\_search}) \\
\mdl{wcaGLM}{GLM}    & \texttt{glm-5.2}                    & thinking                 & first-party search endpoint \\
\mdl{wcaSeed}{Seed}   & \texttt{doubao-seed-2-0-pro}        & reasoning effort high    & native (\texttt{web\_search})   \\
\bottomrule
\end{tabular}
\caption{The six evaluated configurations, as of the end of the tournament. \textbf{Every
model ran at the highest thinking setting its API exposes and used its provider's own
server-side search}; the two columns look heterogeneous only because vendors expose those
controls under different names---a token budget, a reasoning-effort level, or a switch with
no dial. Every retrieval was executed by the provider's own service. Every prediction record
carries a flag attesting whether search was actually exercised; 757 of 848 stored responses
carry it. The cohort is a purposive snapshot: each family's reasoning flagship at the time
of the run.}
\label{tab:models}
\end{table}

\paragraph{Contamination and live evaluation.} Static benchmarks decay as their content
enters pre-training corpora~\citep{sainz2023}, and detection can only bound the damage after
the fact~\citep{timetravel}. The prevailing remedies are continuous
refresh~\citep{livebench,livecodebench} and live human preference~\citep{chatbotarena}. Each
is a proxy for the property one actually wants---that the answer did not exist when the
question was asked. \bench obtains that property directly rather than approximating it, and
pays for it in scope: an event guaranteed uncontaminated is one that happens only once.

\paragraph{Football forecasting and market efficiency.} Statistical models of football
scorelines run from Maher's independent-Poisson formulation~\citep{maher1982} through the
Dixon--Coles correction~\citep{dixon1997} to Bayesian-network
systems~\citep{constantinou2012}. A parallel literature treats bookmaker odds as a forecast
in their own right and finds them hard to beat~\citep{forrest2005}, the football instance of
the efficient-market hypothesis~\citep{fama1970}. Most LLM forecasting work is scored
against chance or against other models; the betting line gives \bench something stronger---a
public, well-calibrated competitor that a model must actually beat rather than merely
exceed.

\paragraph{Relation to \sbench.} \sbench~\citep{societybench} collects Web news and
discourse on a real social event, distils a chronology and generates an audited question
bank along two axes, calibration and temporal accuracy. Because those events have already
occurred, it applies a three-phase entity-and-date anonymisation so that a model cannot
match a story to memory---its single largest assumption. \bench is the same question asked
without that assumption: the tournament had not been played, real names and real dates go to
the models, and ground truth is an objective scoreline. The two are designed to be read as a
pair rather than ranked. \sbench buys breadth---25 events, two languages, two scoring
axes---at the price of trusting anonymisation; \bench buys depth on a single event and needs
no such trust, and so serves as the control condition: if the anonymised setting and the
genuinely-future setting yield the same picture of what these models can and cannot
forecast, the assumption \sbench makes was doing no harm. Together the two bracket the question. \Cref{tab:vs_societybench} states the contrast
term by term.

\section{Method}
\label{sec:method}

The benchmark is one question asked 104 times. What follows is the shape of that question at
a single fixture, the rolling material a model reads before answering it, the three locks
that keep every answer ahead of its own result, and the arithmetic that turns seven fields
into points.

\begin{figure}[t]
\centering
\includegraphics[width=\linewidth]{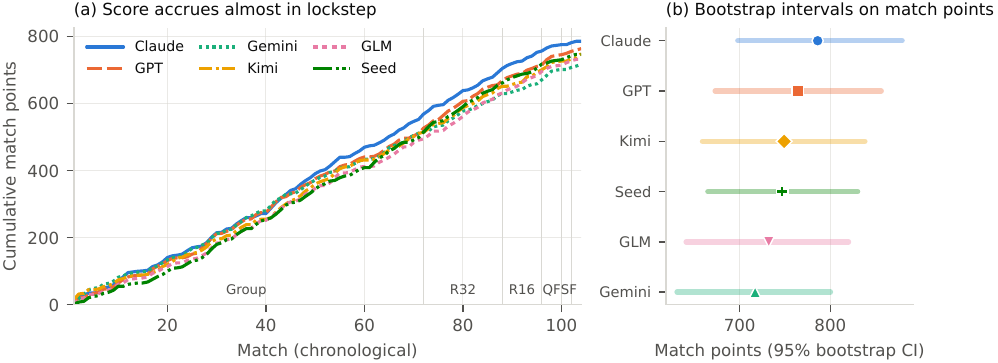}
\caption{(a)~Cumulative match points. The six trajectories are nearly parallel: models
gain and lose together---their herding, seen in the time domain. (b)~Bootstrap intervals on
the match component, alongside the ordering of \cref{tab:leaderboard}.}
\label{fig:leaderboard} \end{figure}

\subsection{Task}
For each match the model receives a header identifying the fixture (competition stage,
group, date, kickoff time in two time zones, venue, city, altitude, home/neutral status,
weather at kickoff, appointed referee, and both head coaches), a fixed bookmaker handicap
line, and the full pre-match dossier of both teams. It is asked to reason step by step and
then emit a single JSON object with exactly seven fields:

\begin{center}\small
\begin{tabular}{@{}llr@{}}
\toprule
\thead
Market & Options & Points \\
\midrule
Handicap (against a fixed line) & home covers / push / away covers & 4 \\
Half-time/full-time             & 9 combinations                   & 3 \\
Match outcome (1X2)             & home / draw / away               & 2 \\
Over/under 2.5 goals            & over / under                     & 2 \\
Both teams to score             & yes / no                         & 2 \\
Correct score                   & \texttt{X--Y}                     & 2 \\
Odd/even total goals            & odd / even                       & 1 \\
\bottomrule
\end{tabular}
\end{center}

\noindent All seven settle on the 90-minute scoreline including stoppage time; extra time
and penalties never count. The handicap line is fixed by us before kickoff and given to
every model, so all six face an identical line. The prompt requires the seven fields to be
mutually consistent with the stated scoreline. The system prompt states that the dossier is
\emph{reference material} and that the model may rely on its own knowledge and searches.
Markets, options and weights were fixed before the tournament and held throughout; the
template wording was refined twice.

\textbf{Every model always sat the same exam.} At each fixture all six models received a
byte-identical prompt---same header, same dossiers, same handicap line, same template---and
answered it within the same window. Every comparison this report draws between models is
therefore a within-fixture comparison under identical conditions, and every refinement of the
template reached all six models alike. What a refinement can affect is comparison
\emph{across time}: a rate computed over early fixtures is comparable to the same rate over
late ones within the same wording. Every prompt sent is archived verbatim.

Two further question types are asked once each, before the tournament: the winner of each of
the 12 groups, fed the four dossiers of that group, and an outright pool---champion, the two
finalists, the four semi-finalists, the winning confederation, and whether total goals
exceed a line of 285.5---fed a compressed 48-team digest. The digest exists for a mundane
reason: the full dossiers run to roughly 420k tokens, far beyond the smallest context in the
cohort, so a fixed extraction compresses them to about 118k. All six receive the identical
digest.

\begin{table}[t]
\centering\small
\setlength{\tabcolsep}{6pt}
\begin{tabular}{@{}lrrrrrr@{}}
\toprule
\thead
Model & Total & Match & Group & Outright & $P(\text{rank 1})$ & 95\% CI (match) \\
\midrule
\mdl{wcaClaude}{Claude} & \textbf{897} & \textbf{786} & 55 & 56 & 0.536 & [698, 879] \\
\mdl{wcaSeed}{Seed}   & 853 & 747 & 50 & 56 & 0.116 & [665, 830] \\
\mdl{wcaGPT}{GPT}    & 845 & 764 & 50 & 31 & 0.206 & [673, 856] \\
\mdl{wcaKimi}{Kimi}   & 839 & 749 & 55 & 35 & 0.104 & [659, 838] \\
\mdl{wcaGemini}{Gemini} & 827 & 717 & 50 & \textbf{60} & 0.012 & [631, 800] \\
\mdl{wcaGLM}{GLM}    & 813 & 732 & 50 & 31 & 0.026 & [641, 820] \\
\bottomrule
\end{tabular}
\caption{Final standings. \emph{Match} is out of a maximum 1664, \emph{Group} out of 60,
\emph{Outright} out of 66. $P(\text{rank 1})$ and the confidence interval come from
10{,}000 bootstrap resamples of the 104 matches.}
\label{tab:leaderboard}
\end{table}

\subsection{Rolling dossiers}
Each team has a Markdown dossier containing a headline block (FIFA ranking, Elo, seeding),
a 14-column squad table with market values, World Cup history, the last four years of
results, key players, recent developments, an overall assessment, and a market-value
section. The frozen pre-tournament baseline totals 623{,}383 characters across the 48
teams; by the final snapshot, after match reports, injury updates and news had been
folded in, it had grown to 1{,}171{,}239 characters. We keep 46 timestamped snapshots, and
each prediction records which snapshot it was served.

\subsection{Three leakage locks}
\label{sec:leakage}
The design is leakage-free by construction, but three procedural locks protect the
construction:

\begin{enumerate}[leftmargin=1.5em,itemsep=2pt,topsep=3pt]
\item \textbf{Snapshot lock.} A prediction is served the snapshot current at that moment,
      which contains results of matches already played and nothing about the match being
      asked. Integration of a match report is gated on the match date being in the past.
\item \textbf{Kickoff lock.} Because the models search the live Web, a prediction issued
      after kickoff could simply retrieve the score. Every prediction in the archive was
      issued before its match kicked off, and the kickoff time is what gates the call.
\item \textbf{Template lock.} The prompt template, market list, scoring weights and the
      handicap line for a match are fixed before the six models are called, and the actual prompt sent is archived verbatim alongside the responses (171 archived prompts), retained for audit.
\end{enumerate}

\subsection{Scoring}

Each market carries the weight shown above; group winners are worth 5
points each and the outright pool 25 (champion), 10 (per correct finalist), 4 (per correct
semi-finalist), 5 (winning confederation) and 4 (total-goals line). A match is therefore
worth at most 16 points and the match component at most $16\times104=1664$. Weights were set
before the tournament and deliberately flattened relative to bookmaker payout intuition,
because several markets that look hard are mostly luck. Scoring is a deterministic script
over two frozen JSON files: the same inputs always give the same totals.

\section{Experiments}
\label{sec:experiments}

That protocol ran once, live, for the length of the tournament. This section first reports
the six systems that sat it and how the run proceeded, then the standings it produced
together with the uncertainty around them, and then two further readings of the same
answers: accuracy market by market against three reference points, and what survives when
the archive is scored five other ways.

\subsection{Setup}
\label{sec:setup}
\Cref{tab:models} lists the six systems. All six are the reasoning-oriented flagship of
their family at the time of the run, and all six were run under the same two rules:
\textbf{thinking turned up as far as the API allows, and search performed by the provider
itself}. Both rules hold for every model in this cohort. What differs between rows of
\cref{tab:models} is vocabulary, not treatment---one vendor takes a token budget, another a
reasoning-effort level, a third exposes a single fixed setting. Retrieval was left entirely
to the provider's own service, which keeps the comparison a test of the model's tool use.
Two models therefore ran on their vendors' first-party endpoints, the only route that
exposes server-side search for them.

The evaluation itself ran for the 39 days of the tournament, from 10~June to 19~July~2026.
Every fixture was answered live, in the hours before its kickoff and in the order the
tournament played them. The archive fixes $104\times7\times6=4{,}368$ match-market picks, $12\times6=72$
group-winner picks and 54 outright slots, for 4{,}494 scored predictions in total; the raw
store holds 848 model responses across 53 batches, comprising 786{,}626 characters of
model-written analysis.

\begin{table}[t]
\centering\small
\setlength{\tabcolsep}{4.5pt}
\begin{tabular}{@{}lrrrrrrr@{}}
\toprule
\thead
 & 1X2 & Handicap & O/U 2.5 & BTTS & Odd/Even & HT-FT & Score \\
\rowcolor{wcaTint} & \textit{(+2)} & \textit{(+4)} & \textit{(+2)} & \textit{(+2)} & \textit{(+1)} & \textit{(+3)} & \textit{(+2)} \\
\midrule
\mdl{wcaClaude}{Claude} & \textbf{66.3} & \textbf{58.7} & 58.7 & \textbf{56.7} & 48.1 & \textbf{35.4} & 14.4 \\
\mdl{wcaGPT}{GPT}    & 63.5 & 52.9 & \textbf{62.5} & 55.8 & 51.9 & 32.9 & \textbf{16.3} \\
\mdl{wcaGemini}{Gemini} & 61.5 & 52.9 & 50.0 & 54.8 & 47.1 & 32.9 & 11.5 \\
\mdl{wcaKimi}{Kimi}   & 64.4 & 53.8 & 54.8 & 53.8 & \textbf{52.9} & 32.9 & 15.4 \\
\mdl{wcaGLM}{GLM}    & 64.4 & 54.8 & 53.8 & 54.8 & 43.3 & 31.6 & 11.5 \\
\mdl{wcaSeed}{Seed}   & 63.5 & 53.8 & 61.5 & 53.8 & \textbf{52.9} & 30.4 & 11.5 \\
\midrule
\textit{Mean}          & 63.9 & 54.5 & 56.9 & 55.0 & 49.4 & 32.7 & 13.5 \\
\textit{Majority vote} & 65.4 & 58.7 & 52.9 & 53.8 & 49.0 & 32.9 & 13.5 \\
\midrule
\rowcolor{wcaTintB}Uniform guess & 33.3 & 50.0 & 50.0 & 50.0 & 50.0 & 11.1 & --- \\
\rowcolor{wcaTintB}Always modal answer & 45.2 & ---  & 53.8 & 54.8 & ---  & ---  & 13.5 \\
\rowcolor{wcaTintB}\textcolor{wcaBase}{\textbf{Bookmaker favourite}} & \textcolor{wcaBase}{\textbf{64.4}} & \textcolor{wcaBase}{\textbf{53.8}} & --- & --- & --- & --- & --- \\
\bottomrule
\end{tabular}
\caption{Per-market accuracy (\%) over 104 matches, with three reference points. The
models clear uniform guessing on 1X2 and HT-FT, but not the bookmaker-favourite baseline
on 1X2, and barely the coin flip on the highest-weighted market, the handicap.}
\label{tab:markets}
\end{table}

\subsection{The leaderboard}
\label{sec:leaderboard}

Claude finishes first with 897 points and GLM last with 813, a spread of 84 points (10.3\%).
\Cref{tab:leaderboard} and \cref{fig:leaderboard}b report what a bootstrap over matches
makes of that spread~\citep{efron1994}. Claude is first in 53.6\% of 10{,}000 resamples; the
paired difference against GPT is $+22$ points with a 95\% interval of $[-49, +94]$, and even
against last-placed-on-match-points Gemini it is $+69$ with $[-6, +148]$. Only the
Claude--Gemini contrast approaches conventional significance ($P(\Delta>0)=0.964$), and it
does so on the component where Gemini is weakest while Gemini simultaneously wins the
outright pool.

Seen over time rather than as a final tally, the order is reasonably stable at both ends.
The cumulative leader changed four times, all of them inside the first twelve days, and
Claude stayed in front from 23~June onward; Gemini and GLM spent most of the second half in
the bottom two. The middle is a different matter---Kimi and Seed trade places from day to
day. \Cref{fig:leaderboard}a plots the trajectories. The margin behind that order is 84
points over 104 matches, and the bootstrap in \cref{tab:leaderboard} gives the leader a 0.54
probability of finishing first on a resampled tournament. Six systems from six
organisations, all at their highest thinking setting and all searching the live Web, land
within a few percentage points of each other on every market we score.

\begin{figure}[t]
\centering
\includegraphics[width=\linewidth]{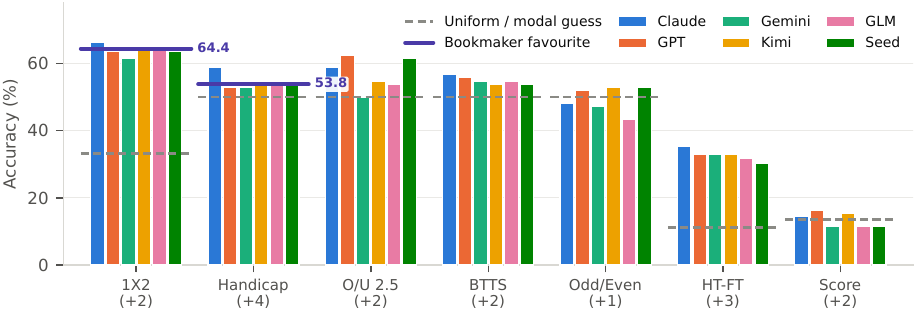}
\caption{Per-market accuracy against uniform/modal guessing (dashed grey) and, where it
applies, the bookmaker-favourite baseline (solid violet). The two violet lines are the
standard these predictions are measured against. Bar colours identify
models consistently across every figure and table.}
\label{fig:markets}
\end{figure}

\subsection{Market-level accuracy}
\label{sec:markets}

\Cref{tab:markets} and \cref{fig:markets} give the market-level picture. The models are far
above uniform guessing where the option set is large---63.9\% against 33.3\% on 1X2, 32.7\%
against 11.1\% on half-time/full-time---and that signal disappears against a competent
baseline. Mechanically backing the side the handicap line favours yields 64.4\% on 1X2,
which only Claude (66.3\%) exceeds, by two matches; on the handicap market itself the
baseline yields 53.8\% against a model range of 52.9--58.7\%. Aggregation does not help
either: a majority vote reaches 65.4\% on 1X2, within noise of the best single model.

\begin{figure}[t]
\centering
\includegraphics[width=\linewidth]{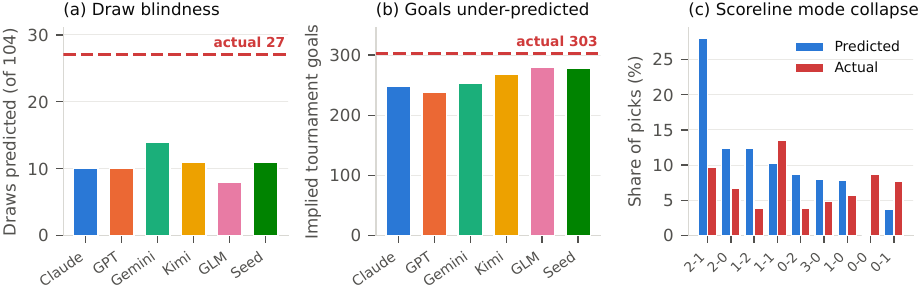}
\caption{Three faces of the same conservatism. (a)~Draws predicted versus the 27 that
occurred. (b)~Total goals implied by the models' scoreline picks versus the 303 actually
scored. (c)~Distribution of scoreline picks against the true distribution; 2--1 absorbs
28.0\% of all picks.}
\label{fig:conservatism}
\end{figure}

\begin{table}[t]
\centering\small
\setlength{\tabcolsep}{5pt}
\begin{tabular}{@{}lrrrrrr@{}}
\toprule
\thead
Round & $n$ & Draws & Model mean & \textcolor{wcaBase}{Bookmaker fav.} & Best model & Points/match \\
\midrule
Group        & 72 & 20 & 64.4 & 63.9 & 68.1 & 7.26 \\
Round of 32  & 16 & 3  & \textbf{86.5} & 81.2 & 87.5 & 8.47 \\
Round of 16  & 8  & 1  & 52.1 & 75.0 & 75.0 & 6.42 \\
Quarter-final& 4  & 2  & 45.8 & 50.0 & 50.0 & 5.79 \\
Semi-final   & 2  & 0  & 8.3  & 0.0  & 50.0 & 4.42 \\
Final \& 3rd & 2  & 1  & 8.3  & 0.0  & 50.0 & 3.75 \\
\bottomrule
\end{tabular}
\caption{1X2 accuracy (\%) and match points by round. Information about the teams is at its
richest in the last four rows and performance at its worst.}
\label{tab:rounds}
\end{table}

\begin{table}[t]
\centering\small
\setlength{\tabcolsep}{5pt}
\begin{tabular}{@{}lccccccr@{}}
\toprule
\thead
Model & Groups (/12) & Champion & Finalists (/2) & Semis (/4) & Confed. & Goals line & Points \\
\midrule
\mdl{wcaClaude}{Claude} & \textbf{11} & \checkmark & 1 & 3 & \checkmark & \checkmark & 56 \\
\mdl{wcaGPT}{GPT}    & 10 & --- & 1 & 3 & \checkmark & \checkmark & 31 \\
\mdl{wcaGemini}{Gemini} & 10 & \checkmark & 1 & \textbf{4} & \checkmark & \checkmark & \textbf{60} \\
\mdl{wcaKimi}{Kimi}   & \textbf{11} & --- & 1 & \textbf{4} & \checkmark & \checkmark & 35 \\
\mdl{wcaGLM}{GLM}    & 10 & --- & 1 & \textbf{4} & \checkmark & --- & 31 \\
\mdl{wcaSeed}{Seed}   & 10 & \checkmark & 1 & 3 & \checkmark & \checkmark & 56 \\
\bottomrule
\end{tabular}
\caption{Tournament-level questions, all answered before the opening match. Ground truth:
champion Spain; finalists Spain and Argentina; semi-finalists France, Spain, England,
Argentina; winning confederation Europe; 303 goals against a line of 285.5.}
\label{tab:outrights}
\end{table}

\Cref{fig:conservatism,tab:rounds,tab:outrights} give the evidence for the five shared properties set out in \cref{sec:intro}: the models track the betting line, under-commit to
draws and to goals, agree with one another far more often than they are right, lose accuracy
as fixtures tighten, and answer tournament-level questions far better than single matches.
Read together, the five describe one property with one cause. Forecasting an unplayed match
sits outside every training objective in this cohort, so what the benchmark
measures is a by-product of general reasoning and retrieval---and by-products of the same
recipe converge. That is the simplest account of why six systems from six organisations
score alike.

\begin{figure}[t]
\centering
\includegraphics[width=\linewidth]{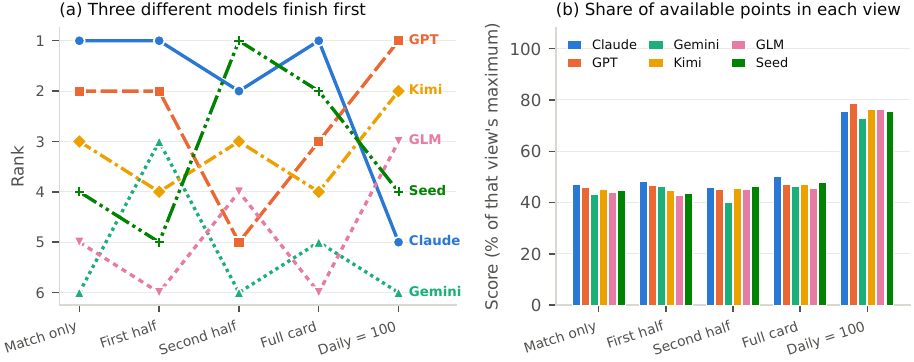}
\caption{(a)~Rank of each model under the five scoring designs of \cref{tab:views}. Three
different models finish first depending only on how the points are added up. (b)~Each view
rescaled to its own maximum, giving every model's share of the points available in it.}
\label{fig:views}
\end{figure}

\subsection{Five scoring designs}
\label{sec:views}

Every number so far rests on one scoring design, and that design is a choice. The archive
lets us test how much the conclusions depend on it: we re-aggregate the same 4{,}494 locked
predictions five ways---the same stored answers throughout, only the arithmetic
changes---and ask whether the ranking survives.

\begin{table}[t]
\centering\small
\setlength{\tabcolsep}{4.2pt}
\begin{tabular}{@{}llrrrrrrl@{}}
\toprule
\thead
\# & View & Claude & GPT & Gemini & Kimi & GLM & Seed & Winner \\
\midrule
1 & Match markets only \textit{(/1664)}       & \textbf{786} & 764 & 717 & 749 & 732 & 747 & Claude \\
2 & First half of tournament \textit{(/832)}  & \textbf{403} & 389 & 384 & 372 & 356 & 361 & Claude \\
3 & Second half of tournament \textit{(/832)} & 383 & 375 & 333 & 377 & 376 & \textbf{386} & \textbf{Seed} \\
4 & Full scorecard \textit{(/1790)}           & \textbf{897} & 845 & 827 & 839 & 813 & 853 & Claude \\
5 & Daily, best model $=100$ \textit{(/100)}  & 75.5 & \textbf{78.7} & 72.8 & 76.3 & 76.2 & 75.7 & \textbf{GPT} \\
\midrule
  & \textit{Rank range across the five} & 1--5 & 1--5 & 3--6 & 2--4 & 3--6 & 1--5 & \\
\bottomrule
\end{tabular}
\caption{The same predictions, five scoring designs. View~1 keeps only the 104 pick cards;
views~2 and~3 split those cards at the chronological midpoint; view~4 is the full scorecard
including group winners and the outright pool; view~5 scores each of the 34 match-days
separately, rescales each day so that day's best model reads 100, and averages the 34
daily indices---which weights days rather than points and so removes the size of a win.
Bold marks the leader of each view.}
\label{tab:views}
\end{table}

\textbf{Three different models finish first.} Claude leads three views, Seed the second half
of the tournament, GPT the daily index; the only ordering statement that survives all five
is that Gemini is never better than third. The two views that overturn the headline each
remove a specific advantage. View~3 splits the tournament at its midpoint---Claude wins the
first half by 14 points and \emph{loses} the second by 3, so its lead is built on
group-stage football. View~5 removes magnitude, counting each day equally, and Claude falls
to fifth. Its advantage is concentrated in a minority of high-scoring days, which view~5 brings out.

\textbf{Averaged over the five, the order is Claude, GPT, Kimi and Seed, GLM, Gemini.} Mean
rank across the designs: 2.0, 2.6, 3.2 and 3.2, 4.8 and 5.2. \Cref{fig:views}b gives each
model's share of the points available under every design.

\begin{figure}[t]
\centering
\includegraphics[width=\linewidth]{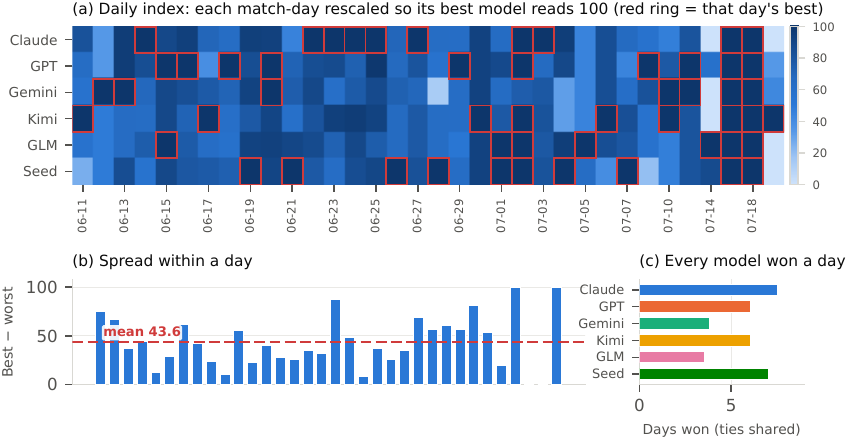}
\caption{The 34 match-days. (a)~Each day rescaled so its best model reads 100; a red ring
marks that day's leader. (b)~The gap between the day's best and worst model. (c)~Days led,
with ties shared.}
\label{fig:days}
\end{figure}

\textbf{Within a day the spread is ten times the tournament-long spread.} The 104 matches fall on 34
days, and a day is what a reader of the live site actually saw (\cref{fig:days}). The mean gap between the best and worst model \emph{within a single day}
is 43.6 points on the 0--100 index, whereas over the whole tournament the six are separated
by 4.1. Resolving a margin of that size scales as $\sqrt{n}$ in the number of matches. Every
model led at least one day, and 26 of the 34 days have a unique leader.

\section{Conclusion}

We ran six frontier LLMs through the entire 2026 World Cup in real time. Each was given a
rolling dossier for all 48 teams, refreshed daily across 46 timestamped snapshots, and was
asked before every kickoff to fill in a seven-market prediction card for the next fixture,
one match at a time, for all 104 matches. The same six also answered 12 group-winner
questions and a pre-tournament outright pool. That is 4{,}494 scored predictions, from 53
prediction batches, alongside 848 stored model responses containing 786{,}626 characters of
model-written analysis and 171 prompts archived byte-identically to what was sent.

Results were settled from the official 90-minute scoreline and written back into the
dossiers only afterwards, so no model ever saw an outcome before committing to it. Because
the tournament had not been played, leakage-freedom is a property of the calendar itself:
the evaluation is clean by construction rather than by filtering.

Scoring is a deterministic script over two frozen JSON files. The benchmark is
released---the 48 dossiers across 46 timestamped snapshots, the fixtures and the official
results---together with the scoring code and the analysis code that regenerates every number
and figure in this report. Prompts, raw model responses and our own predictions are retained
and available for audit; what the public release carries is the question paper and the
answer key. A public site carries the per-match
pick cards, a live leaderboard, and links to the benchmark materials and the pipeline:
\url{https://co-minder.github.io/worldcup2026/}.

\clearpage

\bibliographystyle{abbrvnat}
\bibliography{main}
\end{document}